\documentclass[a4paper]{article}

\usepackage{ISCSLP2024}
\usepackage{tabularx}
\usepackage{multirow}
\usepackage{xcolor}
\usepackage{amsmath}
\usepackage{ifthen}
\newboolean{blind}
\setboolean{blind}{false}

\title{Devising a Set of Compact and Explainable Spoken Language Feature for Screening Alzheimer’s Disease }
\name{
\ifthenelse{\boolean{blind}}{Anonymous to ISCSLP}
{Junan Li$^1$, Yunxiang Li$^1$, Yuren Wang$^2$, Xixin Wu$^1$, Helen Meng$^1$}
}
\address{
  \ifthenelse{\boolean{blind}}{Anonymous to ISCSLP}
  {
  	 $^1$Dept. of Systems Engineering \& Engineering Management, The Chinese University of Hong Kong \\
  $^2$Dept. of Computer Science \& Engineering, The Chinese University of Hong Kong
  }
}

\email{
	\ifthenelse{\boolean{blind}}{Anonymous to ISCSLP}
	{$^1$\{jli,yli,wuxx,hmmeng\}@se.cuhk.edu.hk, $^2$yrwang2@cse.cuhk.edu.hk}
}

\begin{document}

\maketitle

\begin{abstract}
Alzheimer’s disease (AD) has become one of the most significant health challenges in an aging society. The use of spoken language-based AD detection methods has gained prevalence due to their scalability due to their scalability. Based on the Cookie Theft picture description task, we devised an explainable and effective feature set that leverages the visual capabilities of a large language model (LLM) and the Term Frequency-Inverse Document Frequency
(TF-IDF) model . Our experimental results show that the newly proposed features consistently outperform traditional linguistic features across two different classifiers with high dimension efficiency. Our new features can be well explained and interpreted step by step which enhance the interpretability of automatic AD screening.
\end{abstract}

\noindent\textbf{Index Terms}: Alzheimer’s disease detection, spoken language processing, large language models

\section{Introduction}
Alzheimer's disease (AD) detection presents a significant and growing challenge to healthcare and economic systems due to costly and complex diagnoses~\cite{breijyeh2020comprehensive, wimo2007estimate, sh2014review}. Current research underscores the importance of early intervention and the need for economically accessible, non-invasive and affordable alternatives for AD detection~\cite{arvanitakis2019diagnosis, kourtis2019digital, 2015speaking, 9003908}. Consequently, speech and language alternatives have emerged as early AD indicators, offering a promising, non-invasive diagnostic approach suitable for large-scale screening~\cite{calza2021linguistic}. The Cookie Theft picture description task is one of the common cognitive assessment tasks that evaluates language and cognitive impairments through patients' descriptions of a complex scene.

For spoken language-based AD detection, two primary methods have recently been prevalent: linguistic feature extraction with classifier models and the use of pre-trained language models (PLMs) like BERT. Studies such as~\cite{melistascross} focus solely on linguistic features, applying them across multiple languages.~\cite{wagner2023careful} combined linguistic features and classifiers for aphasia subtype classification, incorporating semantic coherence for robust results.

Another approach uses PLMs to capture semantic information and context. For instance,~\cite{li2021comparative} utilized various acoustic and linguistic features, including BERT, to compare different PLMs. They found that BERT-based features significantly improved detection accuracy with both manual and ASR transcripts. Building on this,~\cite{wang2023exploiting} advanced PLMs by incorporating prompt-based fine-tuning for AD detection, aligning training objectives with AD classification tasks for state-of-the-art results. An earlier study by~\cite{li2023leveraging} used PLMs like Whisper and BERT, integrating high-level acoustic and linguistic features along with task-related information to enhance accuracy. More recently,~\cite{heitz2024influence} investigated the impact of using ASR transcription on both linguistic feature-based methods and PLM-based methods.

More recently, Large Language Models (LLMs) such as GPT-4, which have shown remarkable capabilities in various tasks, are increasingly being explored for their potential in aiding AD detection. \cite{Wang2023TextDA} explored the feasibility of using ChatGPT for primary screening of Mild Cognitive Impairment (MCI) based on text conversation analysis. Similarly, \cite{Wang2023CanLL} assessed GPT-4’s potential in dementia diagnosis, highlighting its strengths in zero-shot settings and interpretable explanations, but also its limitations, such as the inability to be fine-tuned and sensitivity to input quality. Moreover, prompt-based LLMs for AD detection face several challenges. Firstly, their outputs are not always controllable or traceable, as changes in the LLMs’ versions may lead to shifts in their outputs. Secondly, substantial computational power is required due to the extremely large size of these models’ parameters. Lastly, there are privacy concerns associated with uploading user data to the cloud.
%Recent studies have highlighted the strong language processing capabilities of LLMs, enabling zero-shot and few-shot learning scenarios that bypass the need for specialized training data \cite{Nori2023CapabilitiesOG}. Due to their powerful language abilities, LLMs have been used in various applications, including AD detection \cite{Lee2023BenefitsLA, Achiam2023GPT4TR}.

Previous research has extensively utilized traditional linguistic features and language models. However, these studies did not explicitly consider task-specific features such as content coverage, which is critical for cognitive assessment. This work introduces a novel set of features, including those that leverage the Term Frequency-Inverse Document Frequency (TF-IDF) concept and features related to the Cookie Theft task. We utilize the advanced linguistic capabilities and visual processsing ability of LLMs to help the generation of our new features. The proposed new features are more interpretable for humans, thereby enhancing the explainability of AD detection.
We compared our new features with 40 traditional linguistic features referenced in the literature \cite{heitz2024influence}. The experimental results demonstrate that our new feature set, which is compact with only around 37.5\% in dimensionality compared with the conventional feature set,consistently outperforms the traditional linguistic features. We achieved an competitive accuracy of 85.4\% on the ADReSS test set using only a 15-dimensional feature set, highlighting the dimensional efficiency of our features.

% To summarize, this work has three main contributions. First, we propose a new, compact and explainable feature set that achieves competitive accuracy in AD detection, demonstrating the effectiveness of our approach. Second, to the best of our knowledge, we are the first to leverage the multimodal abilities of LLMs to generate features for AD by incorporating task-specific knowledge, thus enhancing the depth and interpretability of the diagnostic process. Third, we have pioneered the breakdown and refinement of the Cookie Theft task, making our feature extraction process more traceable and systematic, ensuring that every step is clear and accountable.
% To summarize, this work has three main contributions. First, we have pioneered the breakdown of the Cookie Theft picture and leveraged the visual processing ability of LLM to generate features. Out approach  ensures every step is clear and accountable hence lead to a traceable and explainable feature generation process. Secondly, we use tried and true technique from Information Retrieval (IR) to provide novel features from different aspect. Lastly, we proposed a compact, effective and explainable feature set which achieve a competitive result compared to previous research.
To summarize, this work has three main contributions. First, we pioneered the breakdown of the Cookie Theft picture and leveraged the visual processing ability of LLMs to generate features. Our approach ensures that every step is clear and reasonable, leading to a traceable and explainable feature generation process. Secondly, we utilized tried and true technique from Information Retrieval (IR) to provide novel and grounded features from different perspectives. Lastly, we proposed a compact, effective, and explainable feature set that achieves competitive results compared to previous research.

% This paper is organized as follows: Section \ref{sec:method} details the feature engineering process, including the definition and extraction of the new features. Section \ref{sec:exp} introduces the experimental settings and presents the results. Section \ref{sec:dis} provides a discussion and an ablation study. Finally, Section \ref{sec:con} concludes the paper.
% We also present an in-depth analysis of the results based on these features, offering valuable insights that are likely to benefit future research in AD detection.

%We summarize the major contributions of this paper as follows:
%\begin{enumerate}
%    \item We present several novel explainable features and a mature framework for Alzheimer’s disease detection, which is valuable for future research.
%    \item We did experiments for our new proposed features on various settings, where our features are proven to be useful.
%    \item We analyze the features and got some interesting insights.
%    \item We have made the source code publicly available.
%\end{enumerate}

\section{Method}
\label{sec:method}
\subsection{Dataset}
% In our experiments, we utilize two commonly used dataset, the Pitt corpus from the Dementia Bank database \cite{becker1994natural} and the ADReSS dataset \cite{luz2021alzheimer}. Both focus on the “Cookie Theft” picture description task. The Pitt corpus includes 552 audio recordings from participants describing the Cookie Theft picture, where 309 of them are AD patients and 243 are Healthy Control(HC). The ADReSS dataset, a balanced subset selected from the Pitt corpus, consists of 156 subjects. We select ADReSS dataset mainly for conducting experiments on ASR system generated transcripts and Pitt corpus on manual transcripts.
The dataset utilized in this study is derived from the ADReSS Challenge 2020~\cite{luz2021alzheimer}, which represents a curated subset of the Pitt Corpus within the DementiaBank database~\cite{becker1994natural}. It comprises 156 speech samples and their corresponding transcripts from English-speaking participants engaged in the Cookie Theft picture description task. The participants are categorized into two groups: those without Alzheimer's disease (HC) and those with Alzheimer's disease (AD), with each group including 35 males and 43 females. The dataset is methodically divided into training and testing sets, featuring 108 participants in the training set and 48 participants in the testing set. Both sets are meticulously balanced for age, gender, and disease condition.

\subsection{Feature Engineering}
\begin{table}[t]
  \caption{Fifteen proposed features description}
  \label{tab:des}
  \centering
  \begin{tabularx}{\linewidth}{lX}
    \toprule
    \textbf{Feature Name}      & \textbf{Description}                \\
    \midrule
    Topic 1 Keywords Hit Rate              & Hit rate of the keyword set 1 generated by LLM                        \\
    Topic 2 Keywords Hit Rate                    & Hit rate of the keyword set 2 generated by LLM                              \\
    Topic 3 Keywords Hit Rate          & Hit rate of the keyword set 3 generated by LLM                  \\
    BLEU-1                    & Averaged 1-gram BLEU score with 15 references generated by LLM                                \\
    BLEU-2             &  Averaged 2-gram BLEU score with 15 references generated by LLM \\
    BLEU-3          &  Averaged 3-gram BLEU score with 15 references generated by LLM \\
    BLEU-4          &  Averaged 4-gram BLEU score with 15 references generated by LLM \\
    METEOR			& Averaged METEOR score with 15 references generated by LLM               \\  
    TF-IDF similarity HC        & Cosine similarity with the HC reference vector.                                       \\
    TF-IDF similarity AD        & Cosine similarity with the AD reference vector.                                       \\
    TF-IDF Keywords Hit Rate                   & Hit rate of keywords selected by TF-IDF                                 \\
    avg\_depth                                  & Averaged parse tree depth \\
    Filled Pauses                               & The number of filled pauses \\
    Filled Pauses Ratio                         & The ratio between filled pauses and all tokens \\
    WER                                         & Word error rate
						\\
    \bottomrule
  \end{tabularx}
  \end{table}
In our work, we propose 11 new features for this task. Table~\ref{tab:des} summarizes these 11 features with their corresponding description. 
% \begin{enumerate}
%     \item TF-IDF Score
%     \item Class Importance Score Keywords Hit Rate
%     \item Task related features (8): i-th topic keywords hit rate (3), BLEU (4) and METEOR.
%     % \item LLM-based features (9):  Semantic Repetition, Structural Complexity, Referential Specificity, Language Coherence, Semantic Impairments, Memory Lapses, Linguistic Deficits, Tangentiality and Uncertainty
% \end{enumerate}
In this section, We will introduce the definition and extraction process for each feature in the following section.

\subsubsection{Topic Related Features}
\begin{figure}[!t]
    \centering
    \includegraphics[width=0.5\textwidth]{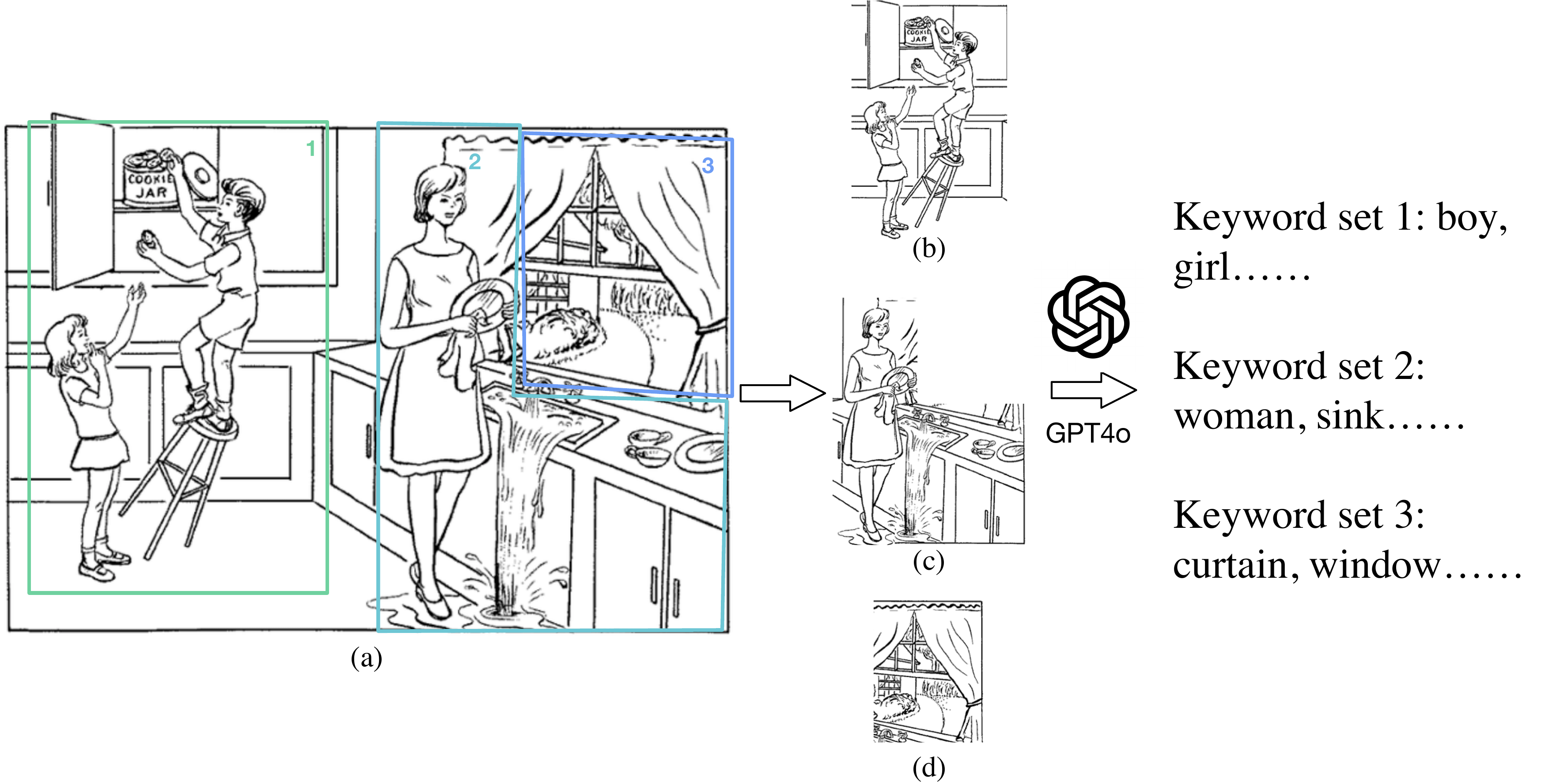}
    \caption{Schematic diagram of topic keyword generation. (a) shows the Cookie Theft picture and how we segment the picture. (b), (c) and (d) show the sub-pictures we crop. Three subpictures are sent to LLM with instructions for generating keywords.}
    \label{fig:cookie-crop}
\end{figure}

A critical aspect of the Cookie Theft picture description task is to evaluate the comprehensiveness of a subject’s description in terms of picture content and topics. As illustrated in Fig.~\ref{fig:cookie-crop}(a), the picture is segmented into three distinct topics: the boy and girl taking cookies, the mother and the water sink, and the window with curtain. 
% We then crop the picture into three sub-pictures according to the topic segmentation. We send the cropped pictures to the multimodal LLM\footnote{The LLM used in this work is the latest version of GPT-4o, 2024-05-13.} and utilize its visual thinking ability to generate related keywords as shown in Fig.\ref{fig:cookie-crop}. For each sub-picture, we generate 50 times and merge the answers to ensure as many as the contents are covered.  With three keyword sets, we determine whether a subject has addressed a topic by checking if their description includes any of the keywords from the respective set. Additionally, we calculate the keyword hit rate within individual topics to quantify description detail degree. The following three types of features are proposed:
We then segment the picture into three sub-images based on the identified topics and send these cropped images to the multimodal LLM\footnote{The LLM used in this work is the latest version of GPT-4o, 2024-05-13.}, leveraging its visual processing capabilities to generate relevant keywords, as illustrated in Fig.~\ref{fig:cookie-crop}. For each sub-picture, we conduct 50 iterations of keyword generation and aggregate the results to ensure comprehensive content coverage. We manually check each iteration's output to prevent any potential hallucination and each step of the generation is trackable. With these three sets of keywords, we calculate the keyword hit rate within each topic to quantify the degree of detail in the descriptions.
% \begin{enumerate}
% 		\item i-th topic keywords hit rate:  The number of mentioned keywords in the i-th keyword set divided by the total number of keywords in the i-th keyword set.
% 		\item topic hit rate: the number of mentioned topic divided by the number of topics.
% 		\item topic vector: A binary vector indicating which topics are mentioned, e.g., ‘101’ means topics 1 and 3 are mentioned.
% \end{enumerate}

% Although the topic keyword method is effective to evaluate the how subject describe a local part of the picture. It still lacks the information from the global picture, such as the connection between topics.To evaluate the the description coverage of the whole picture, we can borrow the metrics from image caption task(e.g. BLEU and METEOR scores). These metrics quantify the match degree between the description and the `golden standard, which is quite suitable for our need. To generate the `golden standard', we input the whole Cookie Theft picture into the multimodal LLM and leverge their visual analysis ability to prompt them generate detailed verbal descriptions for the picture. We ran 15 iteration of the generation and consider the 15 responses as the `golden standards'. The averaged BLEU and METEOR scores over these 15 `golden standards' would be the final score. Towards this end, we propose 5 new features: BLEU-1, BLEU-2, BLEU-3 and BLEU-4 which are calculated by different n-gram scheme, and lastly METEOR scores.
Although the topic keyword method effectively evaluates how a subject describes local parts of the picture, it lacks information from the global picture, such as the connections between topics. To assess the description coverage of the entire picture, we can adopt metrics from the image captioning task (e.g., BLEU and METEOR scores). These metrics quantify the degree of match between the description and the ‘golden standard,’ making them suitable for our needs. To generate the ‘golden standard,’ we input the entire Cookie Theft picture into the multimodal LLM and leverage its visual analysis capability to produce detailed verbal descriptions of the picture. We performed 15 iterations of this generation, considering the 15 responses as the ‘golden standards.’ The averaged BLEU and METEOR scores over these 15 ‘golden standards’ serve as the final score. Accordingly, we propose five new features: BLEU-1, BLEU-2, BLEU-3, and BLEU-4, calculated using different n-gram schemes, and the METEOR score. Table~\ref{tab:prompt} shows the prompt template we used and example response from the LLM.
\begin{table}[t]
  \caption{Prompt template and the response example for topic keywords and description generation.}
  \label{tab:prompt}
  \centering
  \begin{tabularx}{\linewidth}{X}
    \toprule
    \textbf{Instruction for topic keyword generation:}\\
    Imagine you are an expert on cognitive assessment using Cookie Theft picture description task. You have the knowledge of the Cookie Theft picture and the key point to assess the AD. Now I will provide you with a sub-picture of the Cookie Theft picture, please give me some key content words related to that part. These words should be helpful for people to distinguish AD patients that the missing of the words may indicate potential cognitive impairment. Please only give the keywords list separated by comma without any further explanation. \\
    \\
    \textbf{Response example:}\\
    boy, girl, cookie, jar, stool, reaching, cabinet... \\
    \midrule
    \textbf{Instruction for description generation:}\\
    This is the picture of the Cookie Theft description task which is widely used for cognitive assessment. Now imagine that you are an elderly people with healthy cognitive state. Please give me a verbal description of this picture to cover as much content as possible in the picture. \\
    \\
    \textbf{Response example:}\\
    ``Well, the boy has climbed up on a stool to get to the cookie jar in the cupboard. He’s giving a cookie to the girl who’s eagerly waiting for it. The mother is busy washing dishes at the sink but hasn’t noticed that the water is overflowing onto the floor. Outside the window, there are trees and another house, so it’s probably a nice, sunny day.''\\
    \bottomrule
  \end{tabularx}
  \end{table}

\subsubsection{TF-IDF Related Features}
Borrowing the idea of TF-IDF from IR \cite{Manning_Raghavan_Schütze_2008}, we propose a new feature called TF-IDF Score for this task. Let's consider each subject's transcript as a document \( d \) and the training document set as \( D \). Each document \( d \) has a corresponding label (HC or AD). We denote the  HC document set as \( d_{HC} \), AD document set as \( d_{AD} \), where \( d_{HC}, d_{AD} \in D \).
% To calculate the TF-IDF Score for each document (subject), we follow the following steps:
% \begin{enumerate}
%     \item Compute the TF-IDF weight for each term.
%     \item Construct the TF-IDF vector for all documents, where each element corresponds to the TF-IDF weight of a term.
%     \item Average all the TF-IDF vectors in HC calss to get the averged HC TF-IDF vector.
%     \item  Calculate the cosine similarity between each document's TF-IDF vector and the averged HC TF-IDF vector to obtain the TF-IDF Score.
% \end{enumerate}

Then we obtain the TF of the term \( t \) in document \( d \) by calculating the the number of times \( t \) appears in  \( d \) divided by the total number of terms in \( d \)  i.e. 
\begin{equation}
\label{TF}
% \text{TF}(t, d_{HC}) = \frac{f_{t,d_{HC}}}{\sum_{t' \in d_{HC}} f_{t',d_{HC}}}
\text{TF}(t, d) = \frac{f_{t,d}}{\sum_{t' \in d} f_{t',d}}
\end{equation}
where:
\( f_{t,d} \) is the occurrences of term \( t \) in document \( d \). \(\sum_{{t' \in d} f_{t',d}}\) is the total number of terms in \( d \).

% Similarly, we obtain the TF of the term \( t \) in \( d_{A} \) as follows:
% \begin{equation}
% \label{TF}
% \text{TF}(t, d_{A}) = \frac{f_{t,d_{A}}}{\sum_{t' \in d_{A}} f_{t',d_{H}}}
% \end{equation}
% where:
% \( f_{t,d_{H}} \) is total the frequency of term \( t \) in AD document set \( d_{A} \). \(\sum_{t' \in d_{A}} f_{t',d_{A}}\) is the total number of terms in \( d_{A} \).

The inverse document frequency (IDF) is a measure of how much information the word provides, that is, if it is common or rare across all subjects. It is defined as:

\begin{equation}
\text{IDF}(t) = \log \left( \frac{| D |}{|\{d \in D : t \in d\}|} \right)
\end{equation}
where:
\(| D |\) is the total number of documents in the training set \( D \).
\( |\{d \in D : t \in d\}| \) is the number of documents in which the term \( t \) appears (i.e., the document frequency of \( t \)).

Then the TF-IDF weight for a term \( t \) in \( d \) is the product of its TF and IDF:

\begin{equation}
\text{TF-IDF}(t, d) = \text{TF}(t, d) \times \text{IDF}(t)
\end{equation}
% \begin{equation}
% \text{TF-IDF}(t, d_{A}) = \text{TF}(t, d_{A}) \times \text{IDF}(t)
% \end{equation}

Then we construct the TF-IDF vector for each document \( d \in D\). Let \( T \) be the set of unique term from the document set. The i-th value of the vectors coresponds to the i-th term in \( T \)  If the i-th term is in the document the value would be its TF-IDF, else 0, i.e.
% \begin{equation}
% \mathbf{v}_d = [{v}_{d1}, {v}_{d2}, \ldots, {v}_{d|T_{H}|}]
% \end{equation}
% %where \(n = |T|\), the total number of terms in the set \(T\).

% \begin{equation}
% {v}_{di} = 
% \begin{cases} 
% \text{TF-IDF}(t_i, d_{H}) & \text{if } t_i \in d \\ 
% 0 & \text{if } t_i \notin d 
% \end{cases}
% \end{equation}

\begin{equation}
\mathbf{v}_{d} = \left[ 
\begin{cases} 
\text{TF-IDF}(t_i, d) & \text{if } t_i \in d \\ 
0 & \text{if } t_i \notin d 
\end{cases} 
\right]_{i=1}^{|T|}
\end{equation}

where \( |T| \) is the total number of unique terms.

Then the HC reference vector is calculated by averaging the TF-IDF vectors of all documents in the document set \( d_{HC} \):

\begin{equation}
\mathbf{v}_{\text{HC}} = \frac{1}{|d_{HC}|} \sum_{d \in d_{HC}} \mathbf{v}_{d}
\end{equation}

Similarly, by replacing \(d_{HC}\) by \( d_{AD} \) and performing same calculation with Equation (5), we obtain the the AD reference vector \(\mathbf{v}_{\text{AD}}\)

Lastly, two similarity features of \( d \)  are calculated as follows:
\begin{equation}
\text{TF-IDF similarity HC}(d) = \text{CosSimilarity}(\mathbf{v}_{d}, \mathbf{v}_{\text{HC}})
\end{equation}
\begin{equation}
\text{TF-IDF similarity AD}(d) = \text{CosSimilarity}(\mathbf{v}_{d}, \mathbf{v}_{\text{AD}})
\end{equation}
In our analysis, we observed that certain key terms, such as ‘window’ (objects) and ‘overflow’ (actions), may be overlooked by some AD subjects for various reasons. To quantify this observation, we propose using the keyword hit rate as a feature. To select appropriate keywords, we choose the top 30 terms that have the highest values in \( \mathbf{v}_{\text{HC}} \) as keywords. The TF-IDF keyword hit rate is then determined by dividing the number of mentioned keywords by the total number of keywords (30).

We also add four linguistic features that are not included in the previous research into our feature set: averaged parse tree depth, filler pause number, filler pauses ratio and word error rate\footnote{We use Whisper-large-v3 as the ASR system for WER evalaution}.

\section{Experiment}

\begin{table*}[h!]
\centering
\begin{tabular}{cccccc}
\toprule
\textbf{Model} & \textbf{Feature} & \textbf{ACC(\%)} & \textbf{PRE(\%)} & \textbf{REC(\%)} & \textbf{F1(\%)}  \\ 
\midrule
\multirow{3}{*}{RF} & Linguistic Features (40) & 75.0 & 87.5 & 58.3 & 70.0  \\ 
& New Features (15) & \textcolor{red}{\textbf{85.4}} & \textcolor{red}{\textbf{87.0}} & \textcolor{red}{\textbf{83.3}} & \textcolor{red}{\textbf{85.1}}  \\ 
& All Features (55) & 80.6 & 89.3 & 69.6 & 78.2 \\
% & Feature Selection (20) & \textcolor{red}{\textbf{\underline{87.5}}} & \textcolor{red}{\textbf{\underline{100}}} & \textcolor{red}{\textbf{\underline{75.0}}} & \textcolor{red}{\textbf{\underline{85.7}}} \\
\midrule
\multirow{3}{*}{XGBoost} & Linguistic Features (40) & 72.9 & {92.3} & 50.0 & 64.9  \\ 
& New Features (15) & \textbf{\underline{{83.3}}} & {86.4} & \textbf{\underline{{79.2}}} & \textbf{\underline{{82.6}}}  \\ 
& All Features (55) & {79.2} & \textbf{\underline{{93.8}}} & {62.5} & {75.0}\\
% & Feature Selection (20) & \textbf{\underline{83.3}} & {86.4} & \textbf{\underline{79.2}} & \textbf{\underline{82.6}}  \\ 
\bottomrule
\end{tabular}
\caption{The overall results of 3 feature sets on Random Forest and XGBoost. ACC: accuracy, PRE: precision, REC: recall, F1: F1 score.}
\label{table:result}
\end{table*}
\label{sec:exp}
\subsection{Experiment settings}
% We constructed classifiers based on three widely recognized methods: Random Forest (RF), Support Vector Machine (SVM), and XGBoost. To ensure optimal performance, we employed Bayesian Optimization\cite{jones1998efficient} to determine a suitable set of hyperparameters for each model.  Notably, the hyperparameters identified through this process were kept fixed across all settings, ensuring consistency and robustness in our evaluation. We used four different feature sets in our experiment. First one is 40 traditional linguistics features proposed by \cite{heitz2024influence}. Second set is the our 23 new features. We alsao combined linguistic features and new features to be the third feature set. Lastly, we incorporate univariate feature selection method to select top 15 features taht have the highest ANOVA F-value. These 15 features would be the fourth feature set of our experiment.
We constructed classifiers based on two widely recognized methods: Random Forest (RF) and XGBoost. To ensure optimal performance, we employed Bayesian Optimization \cite{jones1998efficient} to determine the appropriate set of hyperparameters for each model. The hyperparameters identified through this process were kept fixed across all settings, ensuring consistency and robustness in our evaluation. In our work, we follow the standard train test split of ADReSS dataset.

We used three different feature sets in our experiment. The first set comprised 40 traditional linguistic features proposed by \cite{heitz2024influence}. The second set included our 15 new features and the third set combined the linguistic features and the new features together. 

% two feature sets, one is 40 traditional linguistics features \cite{heitz2024influence}, and another one is in total 63 features combining 40 traditional linguistic features with our 23 new proposed features.

% To evaluate the performance of our proposed features on manual and Automatic Speech Recognition(ASR) generated transcriptions, we employed a 10-fold cross-validation method, which allows us to assess the model's performance comprehensively. To avoid randomness, we trained and evaluated each setting 100 times,resulting in a total of 1000 training runs per setting. We report both the mean and standard deviation for accuracy and AUC-ROC. AUC-ROC is a popular metric for assessing the discriminative power of predictive models \cite{janssens2020reflection}, as it evaluates performance without needing calibrated predictions or a specific classification threshold. Given our focus on model discrimination rather than calibration, we selected AUC-ROC as our primary metric. 

% In our experiments, we utilized different datasets depending on the experimental setup. We conducted tests on two types of transcripts: manual and ASR-generated. Both types of data are highly valuable to practical scenarios. For manual transcripts, we chose to run our experiments on the Pitt corpus due to its larger dataset, which allowed us to obtain more meaningful and robust results and findings. However, for the ASR-generated transcripts, we focused on the balanced subset of the Pitt corpus known as the Address dataset to save costs.

\subsection{Results}

% Table~\ref{table:result} shows the overall experimental results of this work. It is obvious that our new features consistently outperform the traditional linguistic features on all models. These results demonstrate the effectiveness of the new features. Moreover, these features are more intuitive for human to understand and are more related to the Cookie Theft picture description task, thus enhance the explanablity of spoken language based AD detection
Table~\ref{table:result} presents the overall experimental results of this work. The bold numbers indicate the highest scores within the model and the red numbers represent the best score among all. It is evident that our new features consistently outperform traditional linguistic features. These results highlight the effectiveness of the new features. We achieve the best performance of 85.4\% accuracy, this result is comparable to previous research which uses a fine-tuned BERT model and nearly ten times the number of feature dimensions. Furthermore, the new features are more intuitive for humans to understand and are closely related to the Cookie Theft picture description task, thereby enhancing the explainability of spoken language-based AD detection.

% Also, we found that combining the linguistic features with new features may worsen the performance compared to only use the new features. Thus,  we apply the feature selection to filter noisy features and obtain the best performance of 87.5\% accuracy. This result is comparable to previous research that uses the find-tuned BERT model and nearly ten times of dimensions of features~\cite{li2021comparative}.
We found that combining the linguistic features with the new features may worsen performance compared to using only the new features which suggest the importance of applying feature selection to the linguistic features for filtering some noisy features.

\section{Discussion}
\label{sec:dis}
\begin{figure}[!t]
    \centering
    \includegraphics[width=0.45\textwidth]{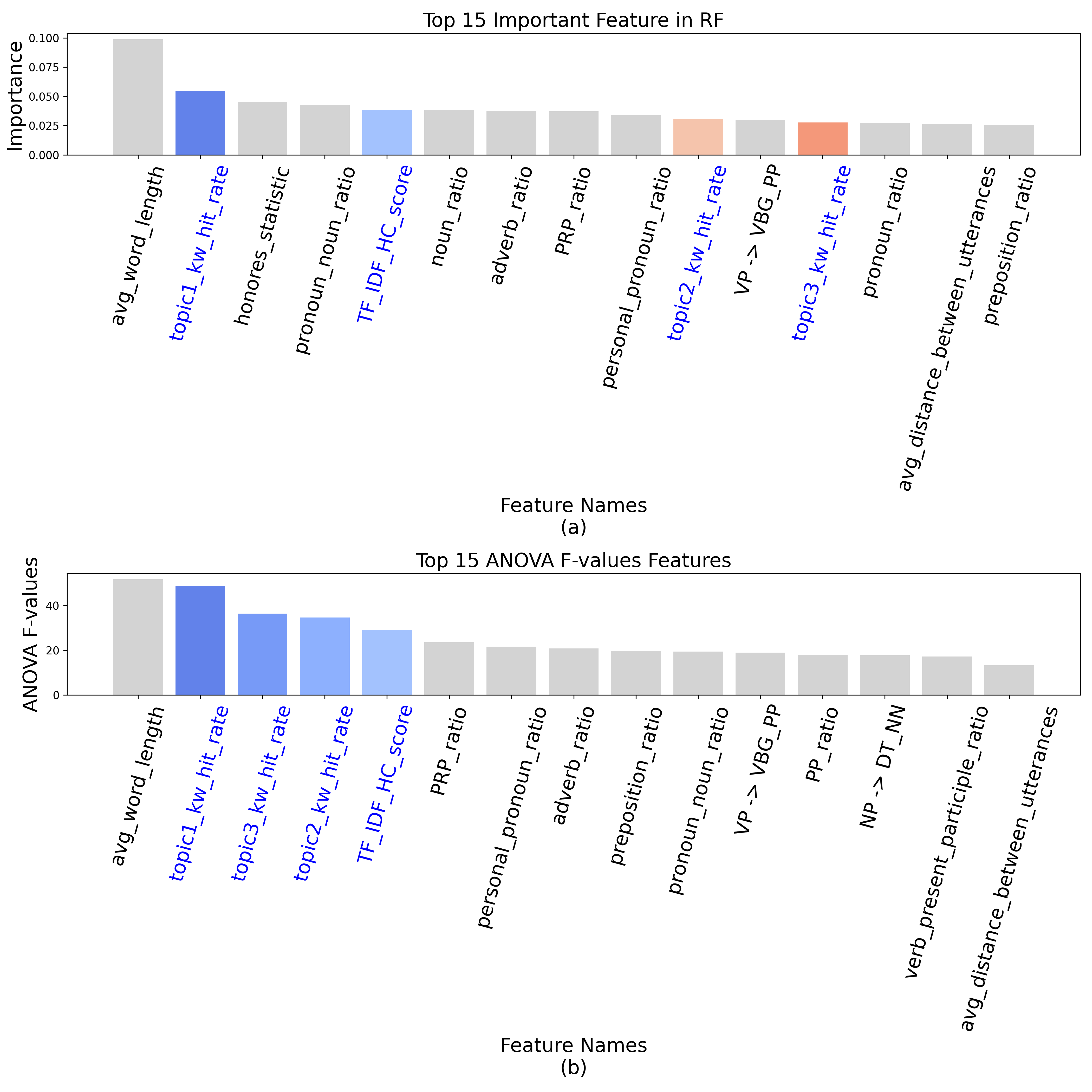}
    \caption{Top 15 Important Features in Random Forest and ANOVA F-values. The charts display the top 15 features ranked by their importance in the Random Forest model (a) and by their ANOVA F-values (b). Newly proposed features  are highlighted in blue names, with their corresponding bars in distinct colors.}
    \label{fig:feat_imp}
\end{figure}
\subsection{Feature Importance and ANOVA F-values}
% To further demonstrate the effectiveness of our features, we extract the feature importance from the RF. Fig.~\ref{fig:feat_imp}(a) shows the top 20 important features in RF. We can see over half (11/20) features are our newly proposed features, especially 4 new features ranked in the top 5. Also, we plot the top 20 features with highest ANOVA F-values. Fig.~\ref{fig:feat_imp}(a) implies that our new features are more relevant to AD than the traditional features that nine of them ranked in the top 10.

% Among our features, we find CIS keyword hit rate and TF\-IDF score are extremly effective which rank at top 3 in both RF feature importance and ANOVA F-values. Moreover, LLM-based features and topic keyword features are also effective, while BLEU and METEOR scores are less important that they donot show up in both figures
To further substantiate the effectiveness of our features, we extracted the feature importance from the RF model. Fig.~\ref{fig:feat_imp}(a) presents the top 15 important features in the RF. Notably, four of our new features ranking in the top fifteen. Additionally, we plotted the top 15 features with the highest ANOVA F-values. Fig.~\ref{fig:feat_imp}(b) indicates that our new features are highly relevant to AD detection as four of them ranking in the top five.

Among our proposed features, we identified that topic 1 keyword hit rate and the TF-IDF similarity HC are particularly effective, consistently ranking in the top five for both the importance of the RF feature and the ANOVA F values. Furthermore, other topic keyword features also demonstrated high effectiveness. 
\subsection{Ablation Study}
\begin{figure}[!t]
    \centering
    \includegraphics[width=0.45\textwidth]{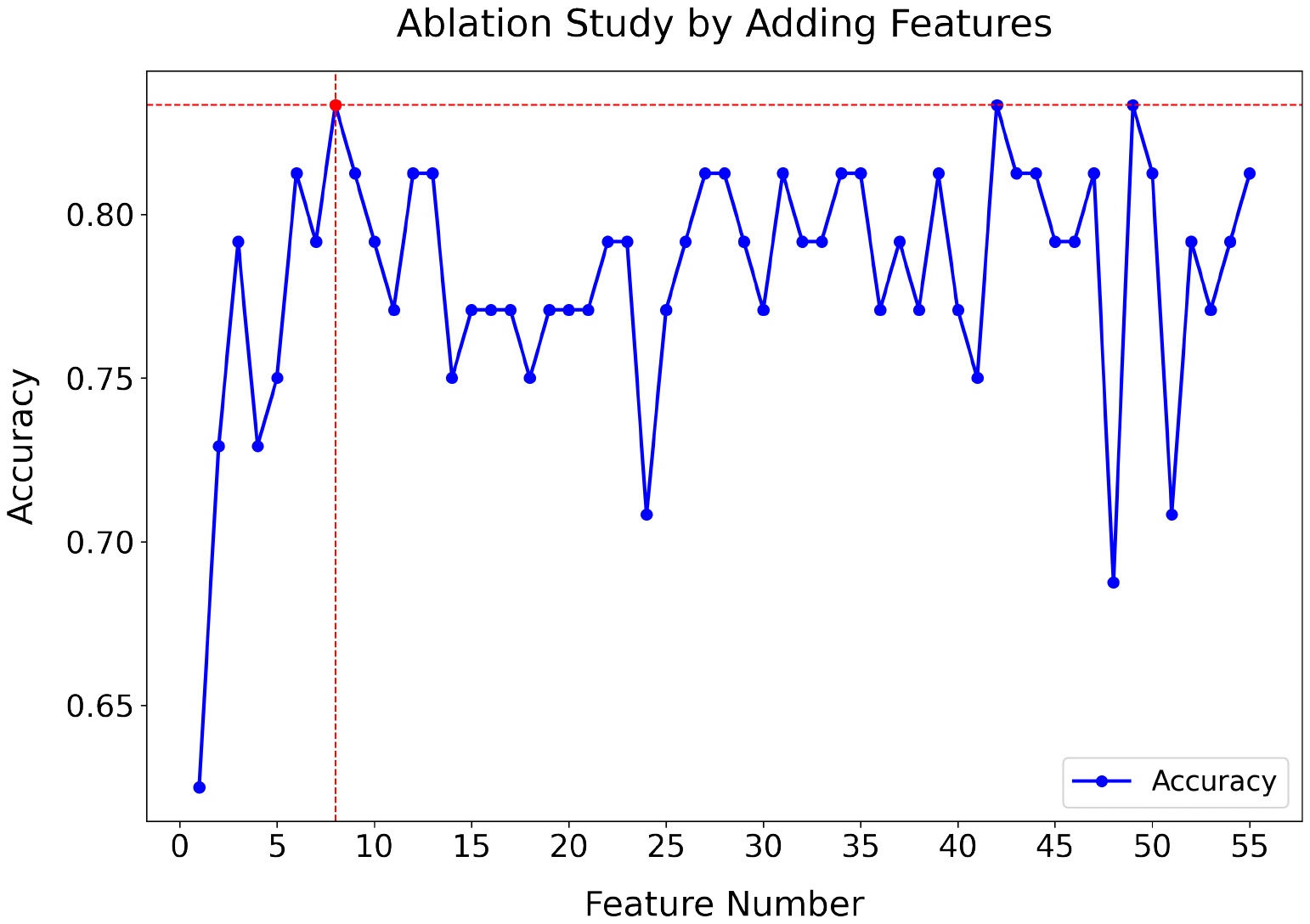}
    \caption{Ablation study accuracy result. The x-axis indicates the number of feature we added into the experiment.}
    \label{fig:ablation}
\end{figure}
% We also presents ablation study for our work. We gradually add features by order of ANOVA F-values and perform AD detection tasks to examine the change in accuracy. Fig~\ref{fig:ablation} presents the results. We can observe a sharp rise of the accuracy from feature number 11 to 15, however, the accuracy immediately fall and vibrate as number of feature goes up. In fact, not all features are contribute positively to the detection performances including some of our new features. We obtained the best result by incorporating 3 linguistic features and 12 new features. Thus, how to organically combine the traditional features with our new features should be one of our next research topics.
We also conducted an ablation study to dive deeper for the inverstigation. We incrementally added features based on their ANOVA F-values and assessed their impact on the accuracy of AD detection tasks. Fig.~\ref{fig:ablation} illustrates the results of this study. A notable increase in accuracy is observed between feature numbers 1 and 8; however, accuracy declines and fluctuates as the number of features increases. 
% This indicates that not all features contribute positively to detection performance, including some of our newly proposed features. 
The optimal result was achieved by incorporating four traditional linguistic features and four new features, however it does not outperform only using new features, hence the feature selection based on ANOVA F-values may not be suitable. Determining a more effective feature selection to better integrate traditional features with our new features will be a focal point for future research.

% We also investigate the effectiveness of LLM-based features by plotting their distribution(see  \ref{sec:appendix}).Three features with distinct distributions between AD patients and HC are highlighted in Figure~\ref{fig:select_GPT}.

% These features demonstrate significant differences in language patterns between the two groups. Firstly, individuals with AD exhibit a markedly higher level of uncertainty in their speech compared to healthy controls according to Fig.~\ref{fig:select_GPT}. Additionally, increased levels of semantic repetition are observed in the AD group that AD patient tend to revisit the previous mentioned topic. Furthermore, differences in referential specificity indicate that individuals with AD may struggle with accurately referring to people, objects, and events, which underscores the cognitive challenges they face.

% These insights underscore the importance of focusing on high-level semantic features when developing predictive models. Future research should aim to corroborate these observations across larger and more diverse datasets to affirm their applicability in clinical diagnostics. Additionally, investigating alternative methods for evaluating these features beyond the use of LLMs presents a valuable avenue for further study.

\section{Conclusion}
\label{sec:con}
In conclusion, we have proposed a compact set of features that are both more explainable and more effective for AD detection. We introduced the concept of leveraging TF-IDF alongside advanced LLMs' viusal processing ability to generate useful features. Our experiments demonstrate that our new features outperform the traditional features and achieve a competitive performance with high dimensional efficiency and interpretability.
% Additionally, we have made many other valuable findings. Several key words and LLM-based features differentiate AD from HC subjects, revealing critical cognitive and linguistic markers.The insights underscore the potential of these features in identifying cognitive impairments associated with AD, highlighting their value in advancing our understanding and diagnostics of the disease.

\section{Acknowledgements}

This work is supported by the HKSARG Research Grants Council’s Theme-based Research Grant Scheme (Project No.T45- 407/19N) and the CUHK Stanley Ho Big Data Decision Research Centre.

\bibliographystyle{IEEEtran}

\bibliography{mybib}

\begin{thebibliography}{10}
\providecommand{\url}[1]{#1}
\csname url@samestyle\endcsname
\providecommand{\newblock}{\relax}
\providecommand{\bibinfo}[2]{#2}
\providecommand{\BIBentrySTDinterwordspacing}{\spaceskip=0pt\relax}
\providecommand{\BIBentryALTinterwordstretchfactor}{4}
\providecommand{\BIBentryALTinterwordspacing}{\spaceskip=\fontdimen2\font plus
\BIBentryALTinterwordstretchfactor\fontdimen3\font minus \fontdimen4\font\relax}
\providecommand{\BIBforeignlanguage}[2]{{%
\expandafter\ifx\csname l@#1\endcsname\relax
\typeout{** WARNING: IEEEtran.bst: No hyphenation pattern has been}%
\typeout{** loaded for the language `#1'. Using the pattern for}%
\typeout{** the default language instead.}%
\else
\language=\csname l@#1\endcsname
\fi
#2}}
\providecommand{\BIBdecl}{\relax}
\BIBdecl

\bibitem{breijyeh2020comprehensive}
Z.~Breijyeh and R.~Karaman, ``Comprehensive review on alzheimer’s disease: causes and treatment,'' \emph{Molecules}, vol.~25, no.~24, p. 5789, 2020.

\bibitem{wimo2007estimate}
A.~Wimo, B.~Winblad, and L.~J{\"o}nsson, ``An estimate of the total worldwide societal costs of dementia in 2005,'' \emph{Alzheimer's \& Dementia}, vol.~3, no.~2, pp. 81--91, 2007.

\bibitem{sh2014review}
V.~L, R.~SH, R.~M, P.~M, L.~J, C.~M, and L.~G., ``Review of brief cognitive tests for patients with suspected dementia,'' \emph{Int {Psychogeriatr}. 2014 {Aug};26(8):1247-62. doi:}, vol. 10., 2014.

\bibitem{arvanitakis2019diagnosis}
Z.~Arvanitakis, R.~C. Shah, and D.~A. Bennett, ``Diagnosis and management of dementia,'' \emph{Jama}, vol. 322, no.~16, pp. 1589--1599, 2019.

\bibitem{kourtis2019digital}
L.~C. Kourtis, O.~B. Regele, J.~M. Wright, and G.~B. Jones, ``Digital biomarkers for alzheimer’s disease: the mobile/wearable devices opportunity,'' \emph{NPJ digital medicine}, vol.~2, no.~1, p.~9, 2019.

\bibitem{2015speaking}
S.~G, H.~I, V.~V, K.~J, and P.~M., ``Speaking in alzheimer's disease,'' \emph{is {That} an {Early} {Sign}? {Importance} of {Changes} in {Language} {Abilities} in {Alzheimer}'s {Disease}. {Front} {Aging} {Neurosci}. 2015 {Oct} 20;7:195. doi:}, vol. 10., 2015.

\bibitem{9003908}
J.~Weiner, C.~Frankenberg, J.~Schröder, and T.~Schultz, ``Speech reveals future risk of developing dementia: Predictive dementia screening from biographic interviews,'' in \emph{2019 IEEE Automatic Speech Recognition and Understanding Workshop (ASRU)}, 2019, pp. 674--681.

\bibitem{calza2021linguistic}
L.~Calz{\`a}, G.~Gagliardi, R.~R. Favretti, and F.~Tamburini, ``Linguistic features and automatic classifiers for identifying mild cognitive impairment and dementia,'' \emph{Computer Speech \& Language}, vol.~65, p. 101113, 2021.

\bibitem{melistascross}
T.~Melistas, L.~Kapelonis, N.~Antoniou, P.~Mitseas, D.~Sgouropoulos, T.~Giannakopoulos, A.~Katsamanis, S.~Narayanan, and N.~Demokritos, ``Cross-lingual features for alzheimer’s dementia detection from speech.''

\bibitem{wagner2023careful}
L.~Wagner, M.~Zusag, and T.~Bloder, ``Careful whisper--leveraging advances in automatic speech recognition for robust and interpretable aphasia subtype classification,'' \emph{arXiv preprint arXiv:2308.01327}, 2023.

\bibitem{li2021comparative}
J.~Li, J.~Yu, Z.~Ye, S.~Wong, M.~Mak, B.~Mak, X.~Liu, and H.~Meng, ``A comparative study of acoustic and linguistic features classification for alzheimer's disease detection,'' in \emph{ICASSP 2021-2021 IEEE International Conference on Acoustics, Speech and Signal Processing (ICASSP)}.\hskip 1em plus 0.5em minus 0.4em\relax IEEE, 2021, pp. 6423--6427.

\bibitem{wang2023exploiting}
Y.~Wang, J.~Deng, T.~Wang, B.~Zheng, S.~Hu, X.~Liu, and H.~Meng, ``Exploiting prompt learning with pre-trained language models for alzheimer’s disease detection,'' in \emph{ICASSP 2023-2023 IEEE International Conference on Acoustics, Speech and Signal Processing (ICASSP)}.\hskip 1em plus 0.5em minus 0.4em\relax IEEE, 2023, pp. 1--5.

\bibitem{li2023leveraging}
J.~Li, K.~Song, J.~Li, B.~Zheng, D.~Li, X.~Wu, X.~Liu, and H.~Meng, ``Leveraging pretrained representations with task-related keywords for alzheimer’s disease detection,'' in \emph{ICASSP 2023-2023 IEEE International Conference on Acoustics, Speech and Signal Processing (ICASSP)}.\hskip 1em plus 0.5em minus 0.4em\relax IEEE, 2023, pp. 1--5.

\bibitem{heitz2024influence}
J.~Heitz, G.~Schneider, and N.~Langer, ``The influence of automatic speech recognition on linguistic features and automatic alzheimer’s disease detection from spontaneous speech,'' in \emph{Proceedings of the 2024 Joint International Conference on Computational Linguistics, Language Resources and Evaluation (LREC-COLING 2024)}, 2024, pp. 15\,955--15\,969.

\bibitem{Wang2023TextDA}
C.~Wang, S.~Liu, A.~Li, and J.~Liu, ``Text dialogue analysis for primary screening of mild cognitive impairment: Development and validation study,'' \emph{Journal of Medical Internet Research}, vol.~25, p. e51501, 2023.

\bibitem{Wang2023CanLL}
Z.~Wang, R.~Li, B.~Dong, J.~Wang, X.~Li, N.~Liu, C.~Mao, W.~Zhang, L.~Dong, J.~Gao \emph{et~al.}, ``Can llms like gpt-4 outperform traditional ai tools in dementia diagnosis? maybe, but not today,'' \emph{arXiv preprint arXiv:2306.01499}, 2023.

\bibitem{luz2021alzheimer}
S.~Luz, F.~Haider, S.~de~la Fuente~Garcia, D.~Fromm, and B.~MacWhinney, ``Alzheimer’s dementia recognition through spontaneous speech,'' \emph{Frontiers in computer science}, vol.~3, p. 780169, 2021.

\bibitem{becker1994natural}
J.~T. Becker, F.~Boiler, O.~L. Lopez, J.~Saxton, and K.~L. McGonigle, ``The natural history of alzheimer's disease: description of study cohort and accuracy of diagnosis,'' \emph{Archives of neurology}, vol.~51, no.~6, pp. 585--594, 1994.

\bibitem{Manning_Raghavan_Schütze_2008}
C.~D. Manning, P.~Raghavan, and H.~Schütze, \emph{Introduction to Information Retrieval}.\hskip 1em plus 0.5em minus 0.4em\relax Cambridge University Press, 2008.

\bibitem{jones1998efficient}
D.~R. Jones, M.~Schonlau, and W.~J. Welch, ``Efficient global optimization of expensive black-box functions,'' \emph{Journal of Global optimization}, vol.~13, pp. 455--492, 1998.

\end{thebibliography}

% \begin{thebibliography}{9}
% \bibitem[1]{Davis80-COP}
%   S.\ B.\ Davis and P.\ Mermelstein,
%   ``Comparison of parametric representation for monosyllabic word recognition in continuously spoken sentences,''
%   \textit{IEEE Transactions on Acoustics, Speech and Signal Processing}, vol.~28, no.~4, pp.~357--366, 1980.
% \bibitem[2]{Rabiner89-ATO}
%   L.\ R.\ Rabiner,
%   ``A tutorial on hidden Markov models and selected applications in speech recognition,''
%   \textit{Proceedings of the IEEE}, vol.~77, no.~2, pp.~257-286, 1989.
% \bibitem[3]{Hastie09-TEO}
%   T.\ Hastie, R.\ Tibshirani, and J.\ Friedman,
%   \textit{The Elements of Statistical Learning -- Data Mining, Inference, and Prediction}.
%   New York: Springer, 2009.
% \bibitem[4]{YourName17-XXX}
%   F.\ Lastname1, F.\ Lastname2, and F.\ Lastname3,
%   ``Title of your ISCSLP 2024 publication,''
%   in \textit{ISCSLP 2024 -- 23\textsuperscript{rd} Annual Conference of the International Speech Communication Association, September 18-22, Incheon, Korea, Proceedings, Proceedings}, 2024, pp.~100--104.
% \end{thebibliography}

\end{document}